\documentclass[lettersize,journal]{IEEEtran}
\usepackage{amsmath,amsfonts}
\usepackage{algorithmic}
\usepackage{algorithm}
\usepackage{array}
\usepackage[caption=false,font=normalsize,labelfont=sf,textfont=sf]{subfig}
\usepackage{textcomp}
\usepackage{stfloats}
\usepackage{url}
\usepackage{verbatim}
\usepackage{graphicx}
\usepackage{cite}

\usepackage[colorlinks,
            linkcolor=black,
            anchorcolor=black,
            citecolor=black]{hyperref}
\usepackage{amsmath,amssymb,amsfonts}
\usepackage{graphicx}
\usepackage{textcomp}
\usepackage{xcolor}
\usepackage{stfloats}
\usepackage{array}
\usepackage{booktabs}

\begin{document}

\title{An Audio-Visual Attention Based Multimodal Network for Fake Talking Face Videos Detection}

\author{Ganglai Wang, Peng Zhang, Lei Xie, Wei Huang, Yufei Zha, Yanning Zhang
\thanks{This paper was produced by the IEEE Publication Technology Group. They are in Piscataway, NJ.}
\thanks{Manuscript received April 19, 2021; revised August 16, 2021.}}



\maketitle

\begin{abstract}
DeepFake based digital facial forgery is threatening the public media security, especially when lip manipulation has been used in talking face generation, the difficulty of fake video detection is further improved. By only changing lip shape to match the given speech, the facial features of identity is hard to be discriminated in such fake talking face videos. Together with the lack of attention on audio stream as the prior knowledge, the detection failure of fake talking face generation also becomes inevitable. Inspired by the decision-making mechanism of human multisensory perception system, which enables the auditory information to enhance post-sensory visual evidence for informed decisions output, in this study, a fake talking face detection framework FTFDNet is proposed by incorporating audio and visual representation to achieve more accurate fake talking face videos detection. Furthermore, an audio-visual attention mechanism (AVAM) is proposed to discover more informative features, which can be seamlessly integrated into any audio-visual CNN architectures by modularization. With the additional AVAM, the proposed FTFDNet is able to achieve a better detection performance on the established dataset (FTFDD). The evaluation of the proposed work has shown an excellent performance on the detection of fake talking face videos, which is able to arrive at a detection rate above 97\%.
\end{abstract}


\vspace{-0.2cm}
\section{Introduction}
Human facial features are unique to everyone, as the symbols of personal identity, they play an important role in social communication. Over the last decades, the human face generation with deep neuron networks has been extensively studied \cite{02, 03, 04, 05, 06, 07}, and those generation methods are uniformly called DeepFake. By regarding the level of manipulation, DeepFake can be usually categorized into four groups \cite{01}: entire face synthesis, identity swap, attribute manipulation and expression swap. Not only limited for the purpose of data augmentation, these unrealistic generated face images may wantonly spread on the Internet and cause a series of moral problems. Fortunately, different fake detection algorithms \cite{08, 09, 10,11,12,13,14,15,16} have been proposed to combat DeepFake, especially for the digital facial forgery.

\begin{figure}[t]
\centering
\centerline{\includegraphics [width=0.45\textwidth]{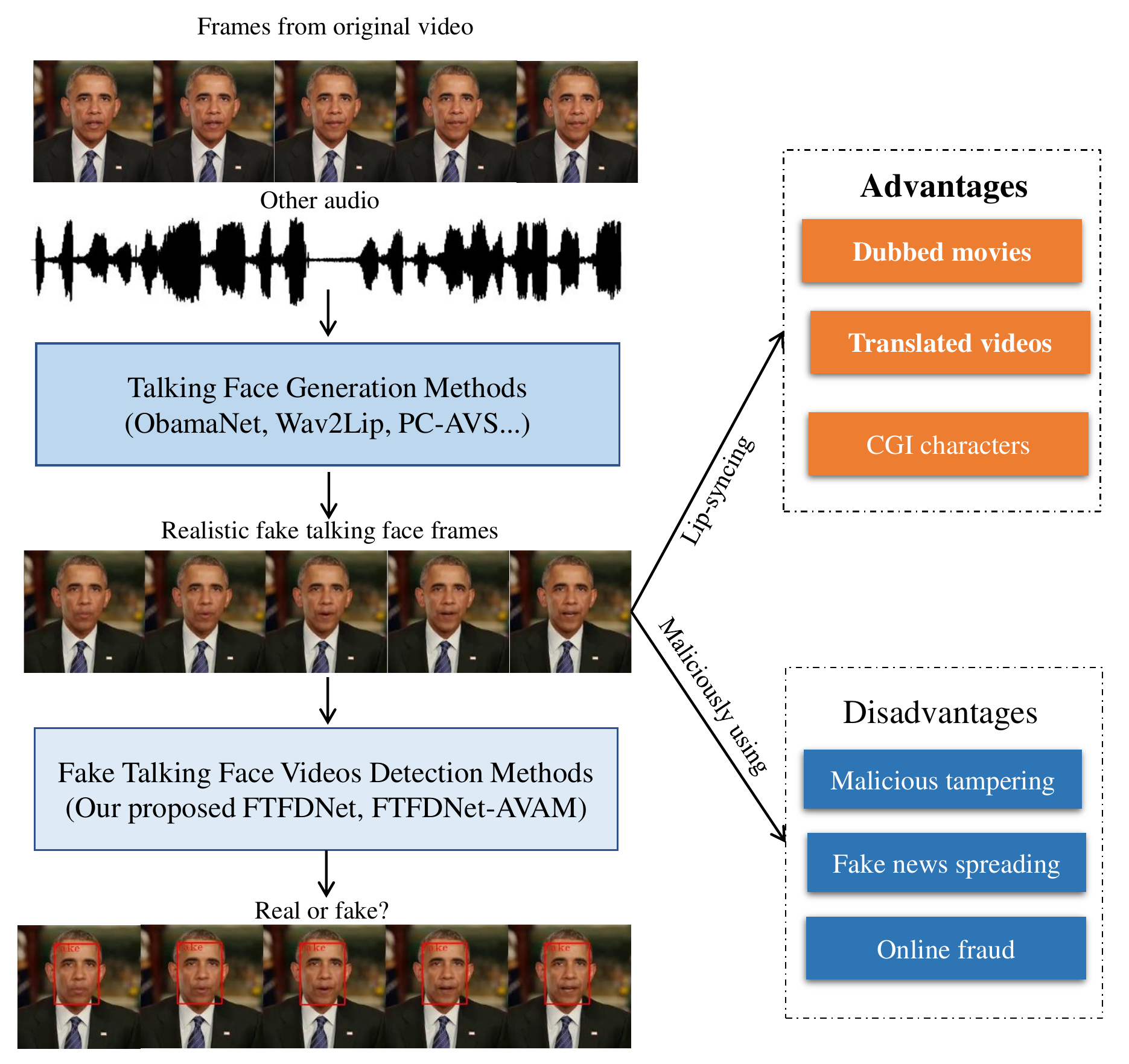}}
\setlength{\abovecaptionskip}{0.cm}
\caption{\textbf{Problem description. }Left: with the help of talking face generation methods, it's easy to generate anyone's talking videos (middle row) using one's pictures and arbitrary speech (top row), our proposed FTFDNet and FTFDNet-AVAM are designed to detect these fake talking face videos (bottom row). Right: the advantages (top row) and disadvantages (bottom row) brought by talking face generation require us to design detection models of fake talking face videos.}
\setlength{\abovecaptionskip}{0.cm}
\label{img_begin}
\vspace{-0.5cm}
\end{figure}

In the age of AI, do you still believe that the voice from a person is realistic or not?  More recently, an emergence of talking face generation has posed a new challenge on fake detection. Compared with common DeepFake, talking face generation only manipulates the lip shape to match the given speech, and does not change facial features of identity, which has stronger concealment. E.g. ObamaNet \cite{17} can synthesis Obama's speech videos based on any text. The work of \cite{18} is also able to generate Obama's talking videos, which only needs a picture of Obama and an arbitrary voice. LipGAN \cite{19} can be used to generate vivid talking faces conditioned on arbitrary identities, voices, and languages. As more and more works having achieved accurate lip synchronization, the generation of undistinguishable videos of fake talking faces is no longer difficult. By using these methods, the face of characters in the video can be easily manipulated, which may help the generated fake talking face videos (e.g., fake news) to spread disinformation and conduct online fraud. Imagine that if there is an email with such a video of assignments from leaders or help from family members, do you believe or not? What you have seen is hard to determined, and `Seeing is Believing` has become a serious challenge.

\begin{figure*}[ht]
\centering
\centerline{\includegraphics [width=1\textwidth]{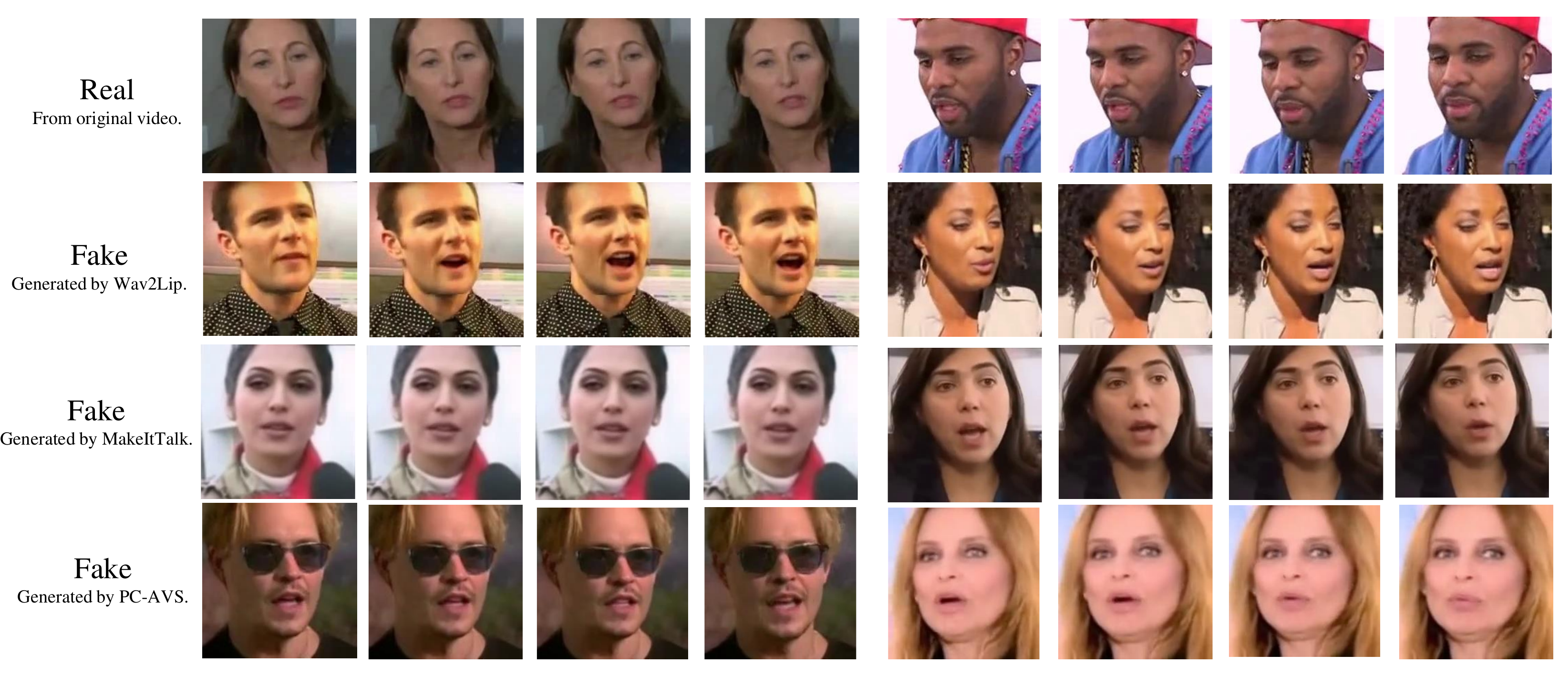}}
\setlength{\abovecaptionskip}{0.cm}
\caption{\textbf{Examples of talking face images from FTFDD. }The images in the first line is from the original videos with out any manipulation. The second line, third line and fourth line are forged by Wav2Lip \cite{26}, MakeItTalk \cite{40} and PC-AVS \cite{28}. All these images are extracted from our established dataset (FTFDD)}
\label{img_visual}
\vspace{-0.5cm}
\end{figure*}

Due to the degradation of frame-wise information in video encoding, it is hard for many fake image detection methods to be directly applied on the DeepFake videos detection. Considering the modeling strategy based on either intra-frame or inter-frame strategies, the proposed DeepFake video detection can be categorized according to the difference of frame-buffering \cite{20, 21, 22, 23}. Elaboratively, the intra-frame methods determine the probability of a fake video by learning a map from only one single video frame, while the inter-frame methods mainly focus on the temporal information between consecutive video frames by building up a buffer of inputs. However, in most of these DeepFake video detection, the audio stream has been totally neglected even it is an essential part of videos. One main reason is that there is no direct correlation between the face forgery and video speech in conventional DeepFake. Instead, talking face forgery is based on the input speech, which usually reshapes the character's mouth appearance to match the given speech. This means that the capability of fake video discrimination may be critically weaken without an effective audio guidance.

According to biological perception mechanism, human multisensory neurons in the superior colliculus is capable of combining multi-modal sensory information about common sources, which is able to improve the ability of object localization and discrimination, even to accelerate the response to them\cite{36,37,38,39,40}. When the visual and audio information is received at the same time, the auditory and visual perception systems are activated to send signals to superior colliculus multisensory neurons, in which the separately obtained information are synthesized as the response to the external stimulation.

With the prior combination of the audio-visual features in a class of neurons, this perceptual mechanism enables the informed decisions output based on a fusion-decoding process. During rapid multisensory decision-making, it has been also verified that the auditory information can enhance post-sensory visual evidence \cite{41}, which inspires a study of fake information perception using cross-modal detection. In this work, we infer the real and fake probabilities by mapping the audio stream and visual modality to the fused low-dimensional features, so as to achieve the detection of fake talking face videos. In addition, a novel audio-visual attention mechanism (AVAM) is proposed to discover which portion of the input feature is more informative to the network. Our AVAM is end-to-end trainable along with CNNs and can be seamlessly embedded into more audio-visual CNN architectures. To achieve a further performance improvement of detection, the AVAM has also been incorporated into our base network for fake talking face detection. The main contributions are summarized as follows:

\begin{itemize}
\item In comparison to DeepFake detection, a more challenging task of fake talking face videos detection is introduced with a proposed visual-audio modeling solution.

\item Inspired by the decision-making mechanism of human multisensory perception system, a novel architecture FTFDNet of informed decisions output is designed based on a fusion-decoding process for fake talking face videos detection.

\item A further improvement of FTFDNet is achieved by incorporating a novel audio-visual attention module to discover more informative features, and the final detection rate arrives at 97\% more.

\item A fake talking face detection dataset (FTFDD) is established on VoxCeleb2 dataset for training and testing.
\end{itemize}

The rest of the paper is organized as follows: Section 2 discussed more details about talking face generation, fake talking face videos detection, and attention mechanism that tightly related to this study. Section 3 introduces the proposed methodology, and Section 4 presents the experiments with discussions, as well as the datasets established for training and testing. In section 5, we conduct ablation studies for more convincible evaluation, and then conclude the whole in Section 6.

\vspace{-0.2cm}
\section{related work}

\noindent \textbf{Talking Face Generation. }To generate a talking face from a given speech is a long-standing matter of great concern in multimedia applications. Kumar \emph{et al.}\cite{17} proposed ObamaNet, in which they employed the Char2Wav architecture (Sotelo et al.\cite{24}) to generate speech from the input text, and then trained the speech synthesis system using the audio and frames extracted from the videos. This approach can be utilized to generate lips with specified identity for any text. Similarly, xxx \emph{et al.}\cite{18} firstly maps the audio features to sparse shape coefficients by RNNs, and followed by mapping from shape to mouth texture/shapes, which finally synthesizes high detailed face textures. \cite{25} proposed Speech2Vid based on a joint embedding of the face and audio to obtain the synthesized talking face video frames with an encoder-decoder convolutional neural network (CNN). The GAN-based LipGAN \cite{19} model inputs the target face with bottom-half masked to act as a pose prior, which guarantees that the generated face crops can seamlessly past back into the original video without further post-processing. As an extension of LipGAN, Wav2Lip \cite{26} employs a pre-trained lip-sync discriminator to correct the lip sync and a visual quality discriminator to improve the visual quality. AttnWav2Lip \cite{27} is a modified version of Wav2Lip obtained by embedding spatial attention and channel attention, which is able to achieve more precise lip-syncing for arbitrary talking face videos with arbitrary speech in the wild. MakeItTalk \cite{40} animates the portrait in a speaker-aware fashion driven by disentangled content and speaker embeddings. \cite{28} proposed Pose-Controllable Audio-Visual System by devising an implicit low-dimension pose code, which generates accurately lip-synced talking faces whose appearances are controllable by other videos.\\

\noindent \textbf{DeepFake Videos \& Fake Talking Face Videos Detection.} With a rapid development with different purpose, it is hard for DeepFake techniques to avoid a dark side, such as generating fake malicious videos of celebrities and masses. To design meaningful work for DeepFake videos detection, \cite{20} pointed out that most image based fake detection can not be directly extended to videos due to a critical degradation of the frames by video compression. Thus, this motivated a facial video forgery detection network (MesoNet) being proposed in \cite{20}, which is integrated by CNNs with a small number of layers. Another fake detection \cite{21} was proposed is obtained by incorporating the attention mechanism into the EfficientNet \cite{29} using Siamese training. Considering the temporal information between consecutive video frames, \cite{22} proposed CLRNet based on Convolutional LSTM and Residual Network to achieve fake detection. Unfortunately, most of those detection strategies failed to take the audio feature into account, whereas it is an essential part of the video. The ignorance of the audio information is because of none-audio-guidance to the traditional face manipulation (eg, entire synthesis,  attribute manipulation ,etc.). But for more advanced fake talking face lip-syncing with given speech reference, the audio-visual representation based modeling would be a promising way to design more effective fake video detectors.

According to the level of manipulation, the DeepFake methods are categorized into 4 main different groups \cite{01}: (1) entire face synthesis: create entire non-existent face images using GAN models. (2) identity swap: replace the face of one person in a video with another one's. (3) attribute manipulation: edit face or retouch face, consists of modifying some facial attributes such as the hair or skin colors, the gender, the age, glass wearing and etc. (4) expression swap: modify the facial expression of the person. In general, all of the DeepFake methods has more or less changes on facial features of identity, but the change of mouth shape usually occurs frequently in videos, and leaves no trace on the person's identity. Different to DeepFake, talking face generation aims at syncing lips to match the input audio, and do not change facial identity features as shown in Fig. \ref{img_visual}. To benefit many applications such as lip-syncing of dubbed movies and news and lectures translation, it also enables people to produce fake talking face videos with malicious purpose, e.g. spreading fake news or extort. And only by the facial features, people has little chance to discriminate whether such a video is true or not, which motivates this study from the existing DeepFake detection.\\


\begin{figure}[htbp]
\centering
\centerline{\includegraphics [width=0.45\textwidth]{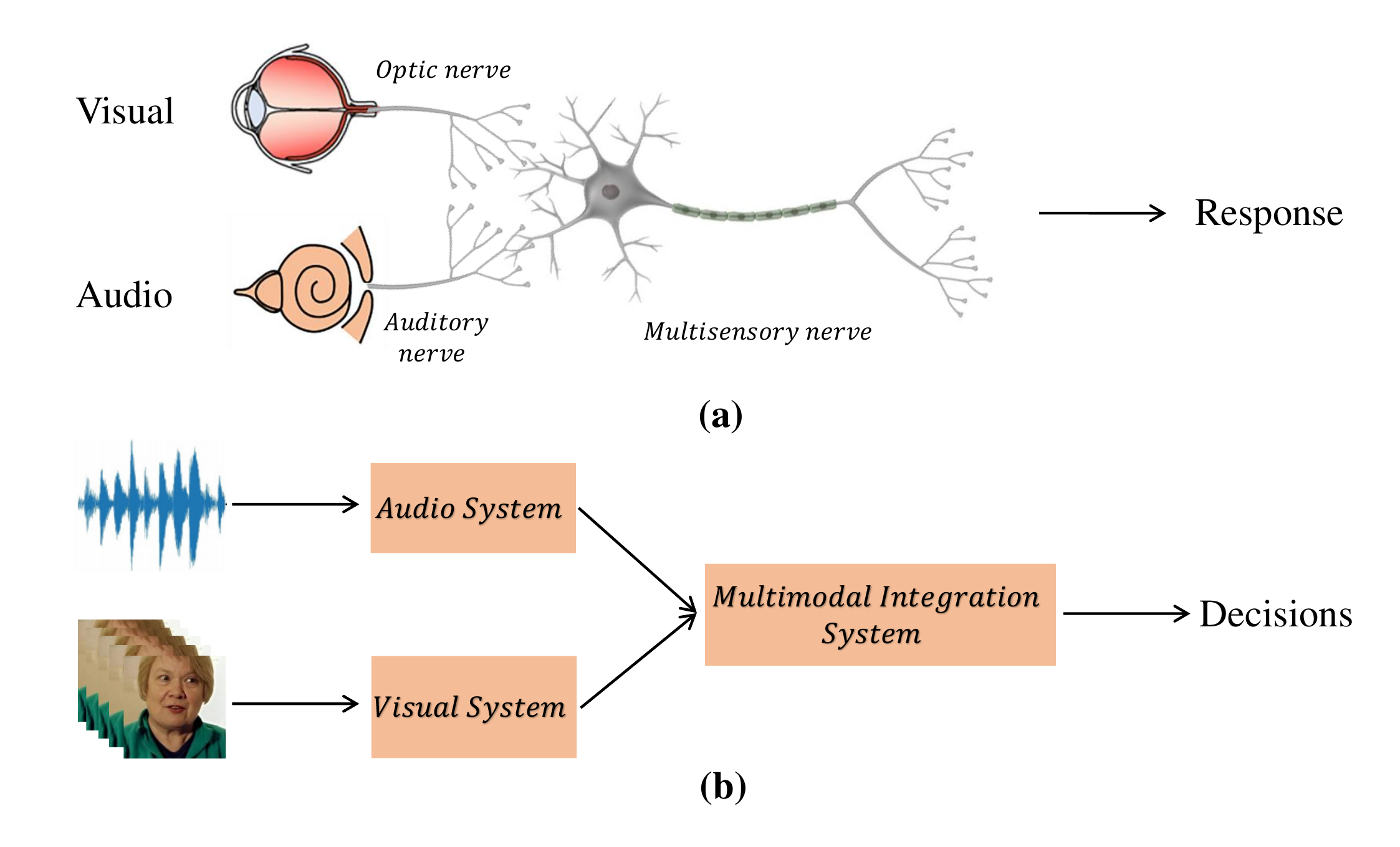}}
\setlength{\abovecaptionskip}{0.cm}
\caption{\textbf{(a) The human multisensory integration system. (b) Human multisensory integration inspired audio-visual decision system. }}
\label{img_human}
\vspace{-0.1cm}
\end{figure}

\noindent \textbf{Human Multisensory Integration System.}The complicated external environment requires the ability more than one sensory modality (five primary senses, vision, touch, hearing, smell, and taste) to be jointly exploited for decision-making. For human, to initiate neuronal response to multimodal environmental events, multisensory neurons in the superior colliculus of midbrain directly integrate signals from different sensory systems(see Fig. \ref{img_human}(a)). This inspires us to generalize this biological perception process of mapping multimodal information to final response as a fusion-decoding process. It is worth noting that the complimentary audio-visual information can also improve our ability to make perceptual decisions compared with visual information only, because the auditory information is capable of enhancing the post-sensory visual evidence \cite{41}. E.g. while crossing the road at night, the vehicle and the surrounding sounds are conveyed to the brain's cerebral cortex after being processed by the visual and auditory system respectively. Then, the separately obtained information is integrated by multisensory neurons, and decoded into sensory response (decision for action) in the end. In this audio-visual process, the sounds from various sources lead to faster accumulation of sensory evidence, which accelerate and correct the final decision-making.\\

\noindent \textbf{Attention Mechanism. }Attention plays an important role in human perception, which enables a person to selectively focus on the salient targets instead of aimlessly processing a whole scenario at one moment. This human biological skill of visual observation can guarantee a high efficiency and accuracy of our visual sensing capability, which has led a wide studies on attention mechanism \cite{46,47,48,49}. Typically, \cite{30} proposed a convolutional block attention module (CBAM) to improve the representative ability of CNNs. By sequentially inferring attention maps along channel and spatial axes, the attention maps are multiplied with the input feature map to obtain the processed maps in which the key regions are effectively emphasized. However, the progress of audio-visual learning, as well as the supplement of audio information to decision-making, also inspired us to design audio-visual attention mechanism, which plays more effective functionality compared to visual attention. Additionally, when the model learning pays more attention to the tampered regions (especially the tiny manipulated region), the proposed audio-visual attention has been validated to achieve forgery detection rather than background processing.

\begin{figure*}[ht]
\centering
\centerline{\includegraphics [width=0.9\textwidth]{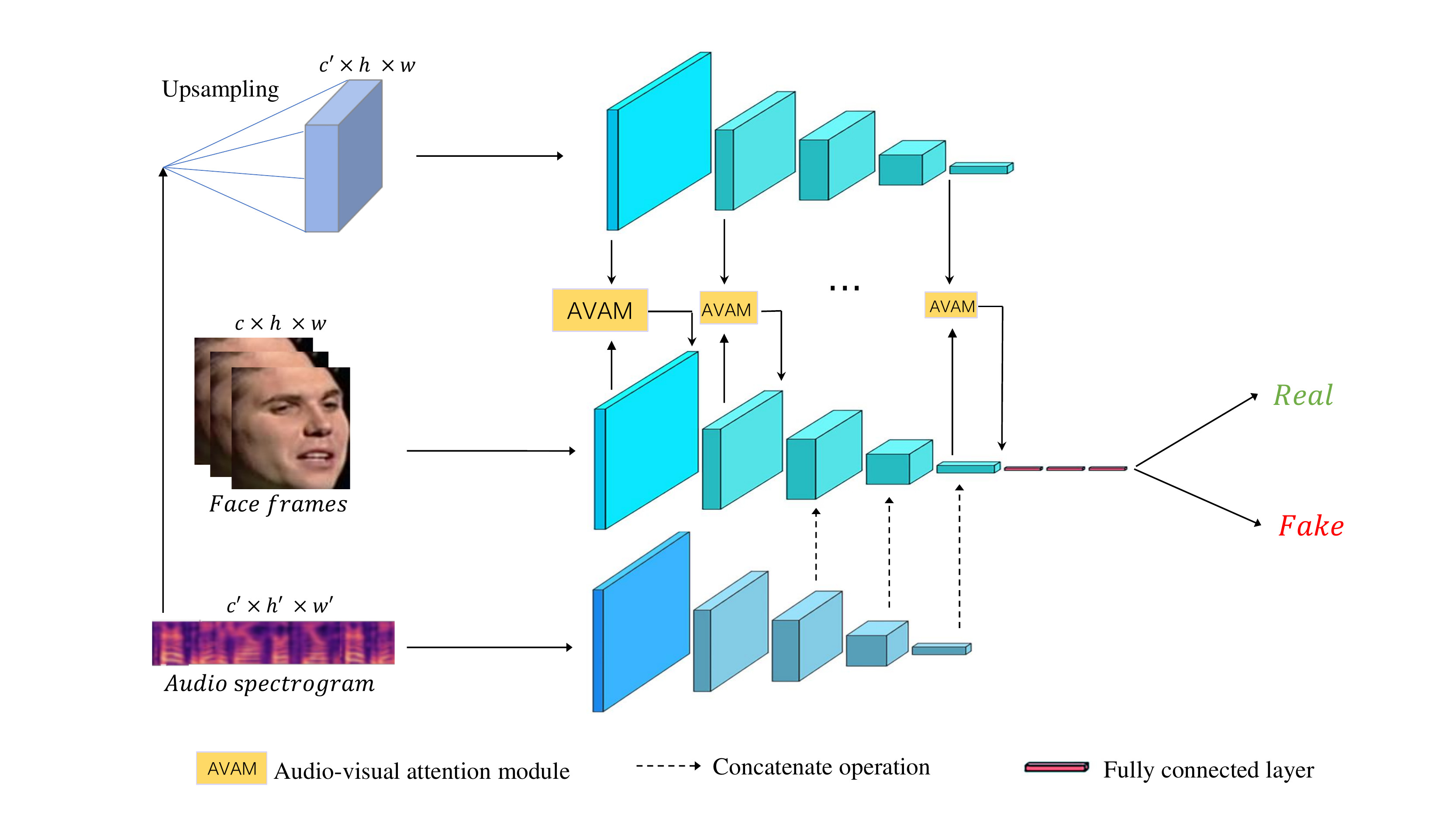}}
\setlength{\abovecaptionskip}{0.cm}
\caption{\textbf{Illustration of our proposed networks. }FTFDNet contains visual branch and audio branch, followed by three fully connected layers. FTFDNet-AVAM, a modified version of FTFDNet is obtained by embedding AVAM into each block of the visual branch.To align the size of audio featrue and visual feature for fusion in AVAM, another network same as visual branch for resized audio feature processing is used.}
\label{img_net}
\vspace{-0.5cm}
\end{figure*}

\vspace{-0.2cm}
\section{proposed methods}
Since the human multisensory system enables us to have a comprehensive feeling of the world by synthesizing the multi-modal senses, especially for visual and audio perceptions, a cross-modal learning strategy was utilized in \cite{41} to recognize multimodal information across different sensory modalities. It also inspires the proposed audio-visual decision scheme as shown in \ref{img_human}(b), which illustrates that the audio helps the visual process to enhance the final decision-making as a supplement. In this section, we firstly introduce our audio-visual fake talking face videos detection network FTFDNet, and followed by the proposed audio-visual attention mechanism AVAM and how to embed it into FTFDNet.
\vspace{-0.2cm}

\subsection{Fake Talking Face Detection Network}
By referring the human multisensory mechanism, the proposed fake talking face detection network is designed with two branches, the visual branch for face encoding and the audio branch for audio stream encoding. The visual branch is composed of a stack of residual convolutional layers and used to encode the consecutive face frames into a low-dimensional feature, while the audio branch is also a stack of residual convolutional layers to encode the input audio spectrogram in accordance with the input faces. The convolution layers employ ReLU activation functions to improve generalization, and Batch Normalization \cite{31} to prevent the vanishing gradient effect by regularizing output. To achieve an audio-visual joint representation, we concatenate feature maps generated by the last three stages of the two branches (both contain five stages) along channel axis.

Following the audio-visual representation, the proposed FTFDNet is composed with three fully connected layers to map the output of joint process into the probability classified as real or fake. By employing Dropout \cite{32} to regularize and improve robustness, FTFDNet is trained to minimize binary cross entropy loss between the predictive probability $y$ and the target label $\hat y$ as Equation \ref{L_{BCE}}. The architecture of FTFDNet is showed on Fig. \ref{img_net}.

\vspace{-0.43cm}
\begin{equation}
L_{BCE} = -\frac{1}{N} \sum_{i=1}^{N}{[y_i \times log(S(\hat y_i)) + [1 - y_i] \times log(1 - S(\hat y_i))]}
\label{L_{BCE}}
\end{equation}

where $\hat y_i$ represents the predictive $i$-th face score, $y_i \in \lbrace 0,1\rbrace$ is the target face label, and $0$ is associated with faces coming from real pristine videos and label $1$ with fake videos. $N$ is the total number of face frames used for training and $S (\cdot)$ is the Sigmoid function.

\begin{figure*}[ht]
\centering
\centerline{\includegraphics [width=0.9\textwidth]{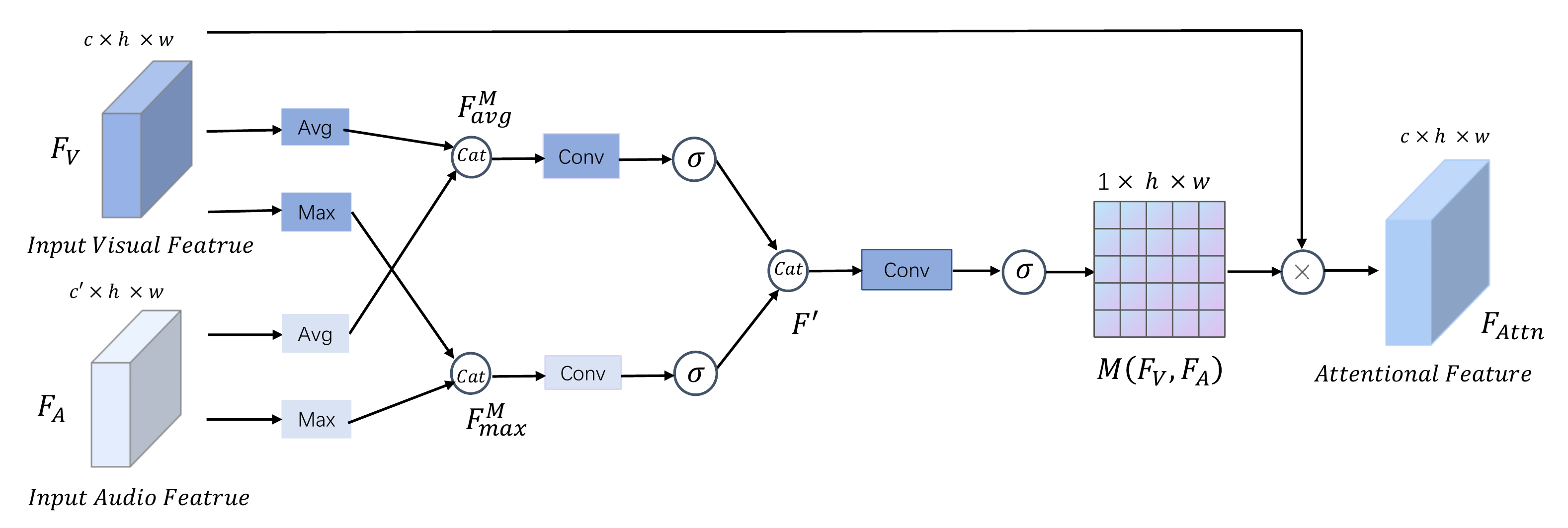}}
\setlength{\abovecaptionskip}{0.cm}
\caption{ \textbf{Illustration of AVAM. }Campared with conventional visual attention mechanism, especially CBAM, the audio information as a supplement are fused into attention maps in a similar way like the process of visual features.}
\label{img_attn}
\vspace{-0.5cm}
\end{figure*}

\vspace{-0.2cm}
\subsection{Audio-Visual Attention Mechanism}
Different from conventional convolutional neural network to treat each position in the whole image equally, the vision tasks require the attention operation to be focused only on the specific regions. For this purpose, the convolutional block attention module (CBAM) \cite{30} has been widely used by discovering which portion in the image can greatly benefit the networks more information. But for audio-visual tasks, the lack of audio information would inevitably weak the information sufficiency of attention mechanisms, and this motivated us to design an more effective audio-visual attention module (AVAM) as below:

%
%
%

Given an intermediate visual feature map $F_V \in \mathbb{R}^{C_1 \times H \times W }$ and a corresponding audio feature map $F_A \in \mathbb{R}^{C_2 \times H \times W}$ as input, AVAM infers a 2D attention map $M \in \mathbb{R}^{1 \times H \times W}$ as illustrated in Fig. \ref{img_attn}. In the beginning, an average-pooling operation is applied along the channel axis to both $F_v$ and $F_A$, which are concatenated to generate an feature descriptor $F_{avg}^M  \in \mathbb{R}^{2 \times H \times W}$. In the same way, $F_{max}^M \in \mathbb{R}^{2 \times H \times W}$ is obtained by utilizing max-pooling operation. Then, we apply a convolution layer followed by a sigmoid activation function to $F_{avg}^M$ and $F_{max}^M$ separately, and concatenate them to generate an intermediate feature map  $F^{\prime} \in \mathbb{R}^{2 \times H \times W}$. Finally, another convolution layer followed by a sigmoid activation function is utilized to generate an attention map $M(F_V, F_A) \in \mathbb{R}^{H \times W}$, which encodes the regions to be emphasized or suppressed. The audio-visual attention map is computed as Equation \ref{M}.

\vspace{-0.1cm}
\begin{equation}
\begin{aligned}
& F_{avg}^M = Concat(AvgPool(F_V), AvgPool(F_A)) \\
& F_{max}^M = Concat(MaxPool(F_V), MaxPool(F_A)) \\
& F^{\prime} = Concat(\sigma(Conv(F_{avg}^M)), \sigma(Conv(F_{max}^M))) \\
& M(F_V, F_A) = \sigma(Conv(F^{\prime}))
\label{M}
\end{aligned}
\end{equation}

where $AvgPool$ and $AvgPool$ denote the average-pooling and max-pooling operations, respectively. $concat$ is the concatenation operation, $Conv$ represents a convolution operation with the filter size of $7 \times 7$, and $\sigma$ denotes the sigmoid function.

With above operations, the attention map $M(F_V, F_A) \in \mathbb{R}^{H \times W}$ is multiplied with the input visual feature map for adaptive feature refinement as described in Eq. \ref{AVAM}. In the new visual feature map $F_{Attn} \in \mathbb{R}^{C_1 \times H \times W}$, the informative regions are emphasized and the inessential parts are suppressed as well. How to incorporate the AVAM into the proposed fake talking face detection network is presented in detail as below.

\vspace{-0.1cm}
\begin{equation}
F_{Attn} = F_V \,\otimes\, M(F_V, F_A)
\label{AVAM}
\end{equation}

where $\otimes$ denote element-wise multiplication.

\begin{figure*}[ht]
\centering
\centerline{\includegraphics [width=1\textwidth]{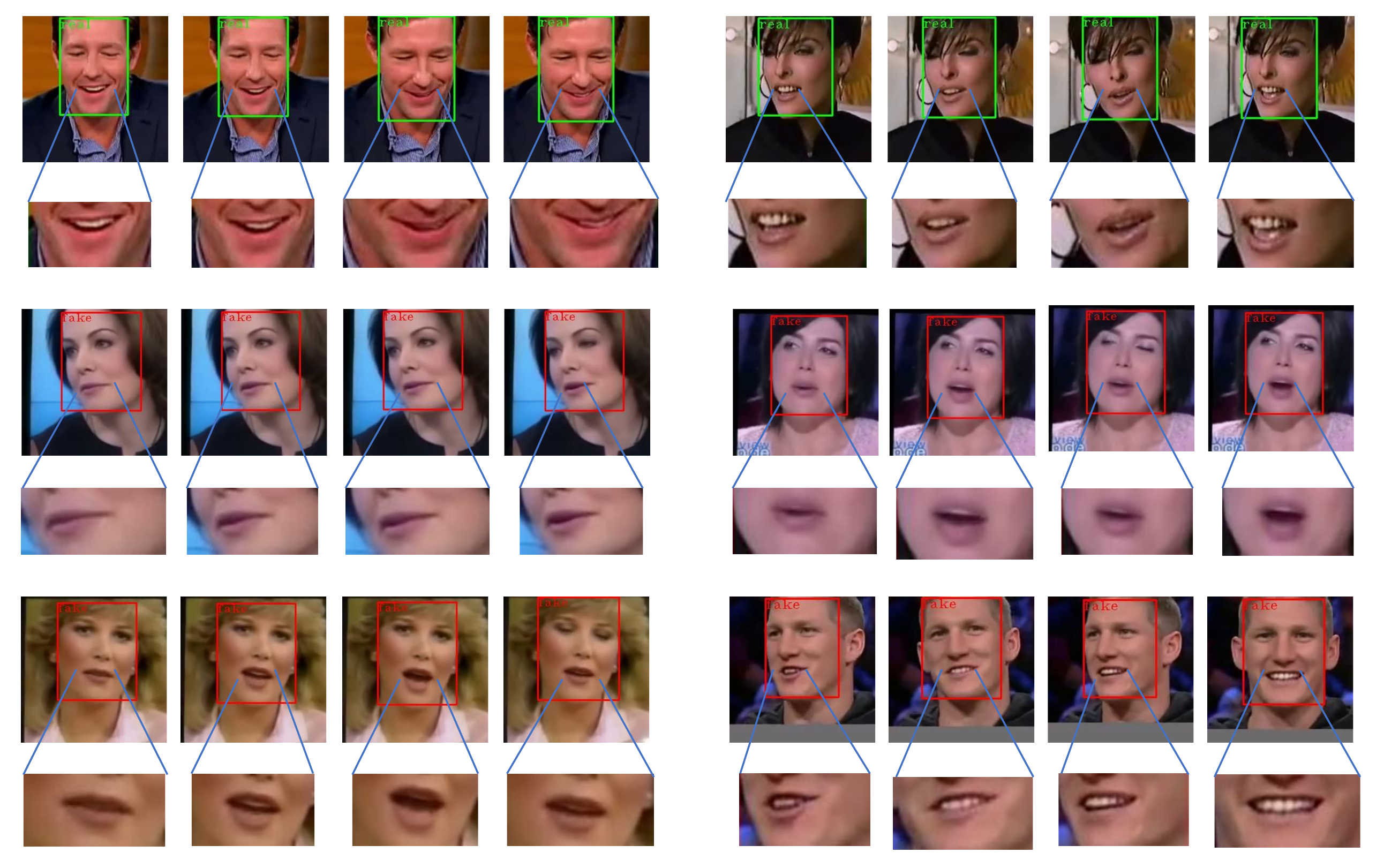}}
\setlength{\abovecaptionskip}{0.cm}
\caption{\textbf{Examples of deepfake talking faces detection. }First line: the original talking faces from video. Next two lines: deep fake talking faces from test set on FTFDD which are difficult to be recognized with the naked eyes. Our model can precisely identify whether any given video is real or fake.}
\label{detection_results}
\vspace{-0.5cm}
\end{figure*}


In talking face generation, the tampering usually happens in limited regions of the whole face, which indicates that detection network should pay more attention to these informative regions, e.g. the lip region for matching the input audio, cheek regions for coordinated vision. Considering the guidance of audio to lip shape, the audio-visual attention module instead of conventional visual attention module is integrated with the FTFDNet.

%
%
%
%

For the fusion of audio and visual features in AVAM, it is critical to ensure their dimensions (width $w$ and height $h$) to be the same. Unfortunately, the initial MFCC features of the input audio and image are usually dimensional mismatching due to the mode difference. In addition, such mismatching also exists among the sizes of the intermediate feature maps caused by divergent functionalities between CNN blocks. In our work, the spectrogram of input audio stream is first resized to the same width and height as the input visual frames. Then, a Siamese like visual branch of FTFDNet is utilized to generate the intermediate audio feature map $F_A$, which is corresponding to the visual feature map $F_V$ generated by CNN blocks in the visual branch. By replacing the original visual feature map $F_A$ with the newly attentional feature map $F_{Attn}$ computed by Eq.\ref{M} and Eq.\ref{AVAM}, an audio-visual attention based fake talking face detection network (FTFDNet-AVAM) is formed. To achieve an overall network optimization, the AVAM is integrated into each block of the visual branch (5 blocks) in FTFDNet.

\section{experiments}
In this section, we report all the details about the established dataset, experiment configurations, results as well as the analysis.
\vspace{-0.2cm}
\subsection{Fake Talking Face Detection Dataset}

Even the DeepFake video detection has attracted a lot of attention recently, to our limited knowledge, the fake talking face video detection has not been widely studied yet. This means that the datasets of DeepFake talking face videos for training and testing is still far from sufficiency for the public. Fortunately, a series of talking face generation approaches \cite{17, 18, 19, 24, 25, 26, 27, 28, 40}  make it possible for dataset generation, which can be categorized into three types:

\begin{itemize}
\item Constrained talking face generation from speech: only can generate talking faces of designated person with a language, which is caused by the model training on a limited dataset of specific person's presentation \cite{14, 17, 18}.

\item Unconstrained speech-driven talking face generation with lip motions only: can generate talking faces of arbitrary identities, voices, and languages \cite{25, 19, 26}.

\item Unconstrained speech-driven talking face generation with lip motions, head poses, facial expression, etc: can generate person's talking face with other facial features in addition to lip movement \cite{40, 28, 51}.

\end{itemize}

In our experiment, high performance models, e.g. Wav2Lip \cite{26}, MakeItTalk \cite{40}, PC-AVS \cite{28} , etc, have been employed to generate fake talking face videos on VoxCeleb2, which is an audio-visual dataset consisting of short clips of human speech, extracted from interview videos uploaded to YouTube. The establishment of fake talking face detection dataset (FTFDD) is introduced in detail as below.

%
%

\begin{figure}[tp]
\centering
\centerline{\includegraphics [width=0.3\textwidth]{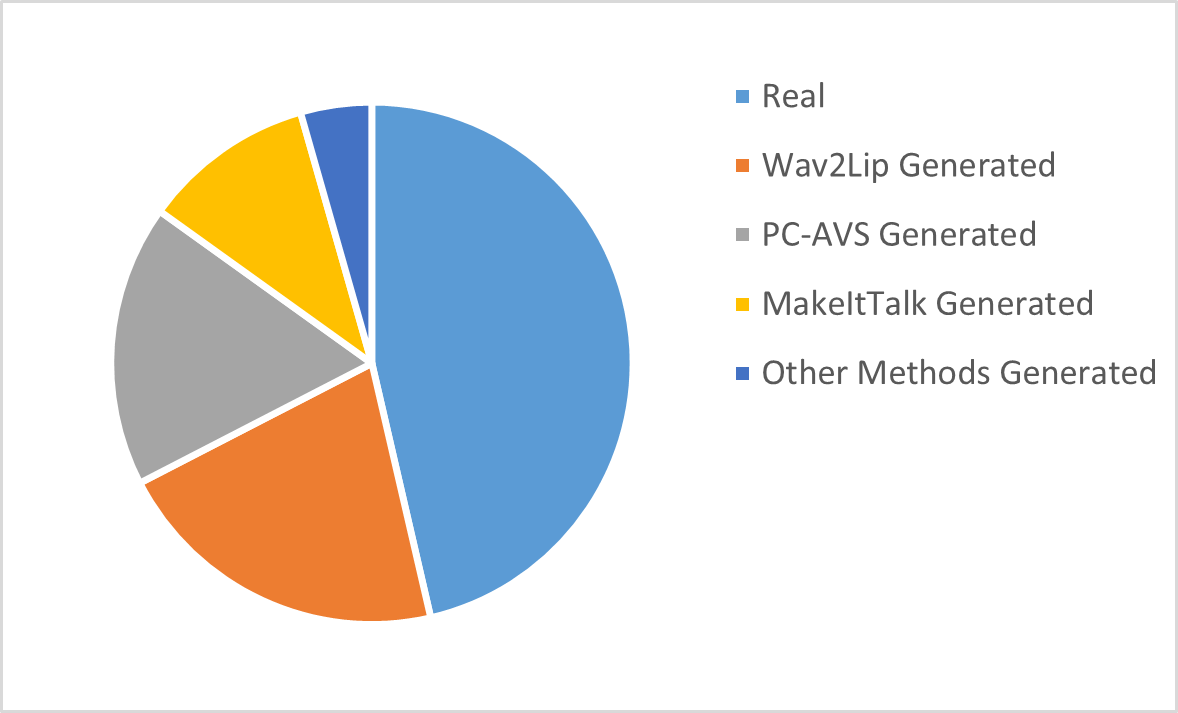}}
\setlength{\abovecaptionskip}{0.cm}
\caption{\textbf{The quantity distribution map of videos generated by different methods. }Fake talking face videos on FTFDD were mainly generated by Wav2Lip \cite{26}, MakeItTalk \cite{40}, PC-AVS \cite{28}, which are  high-performance methods of talking face generation.}
\label{Pie Chart}
\vspace{-0.2cm}
\end{figure}

There are more than 7000 celebrities in VoxCeleb2 covering over 1 million utterances. Considering that each person's video in the data set has many (repeatability), only 70000 more videos are selected as an alternative any of them are randomly selected from a person's utterances. 30000 videos are chosen from the alternative as positive samples, and the rest are used to generate fake talking videos. For each video used to generate fake video, an audio stream is taken from another randomly-sampled video in the alternative. It is noteworthy that some methods of talking face generation only require a single portrait image as an identity reference (MakeItTalk and PC-AVS), which means that the first frame of the video is selected by default as the portrait image. Specifically, there is an extra video required by the PC-AVS as head pose reference, and we used the original video as the pose video reference to obtain more harmonious visions. By removing the videos of generation failure or no face detected while synthesizing, more than 30000 fake talking face videos are left. Together with 30000 genuine videos discussed above, there are about 60000 videos (each segment is at least 1.6 seconds long) are included in our fake talking face detection dataset (FTFDD). The duration of FTFDD is more than 120 hours in total, and 60\% of data are used for model training, 20\% for model validation, 20\% for well-trained model testing. Some sample faces in FTFDD is shown in Fig. \ref{img_visual}, the accurate number of the videos in each class of the FTFDD can be found in Table \ref{table_dataset}. The quantity distribution map of videos generated by different methods is also shown in Fig. \ref{Pie Chart}.

\begin{table}[htbp]
\centering
\caption{\textbf{Cardinality of each class in the created datasets.}The duration and number of real, fake and total videos on FTFDD are shown in this table.}
\begin{tabular}{lccc}
\toprule[1.5pt]
\rule{0pt}{6pt}
 & \text { Fake } & \text { Real } & \text { Total }\\
\hline
\rule{0pt}{8pt}
\text { Total Duration(h) } & 62.28 & 61.05 & 123.33\\
\rule{0pt}{8pt}
\text { Number } & 34679 & 30000 & 64679\\
\text { Minimum Duration(s) } & 1.68 & 1.80 & 1.68\\
\text { Maximum Duration(s) } & 122.26 & 133.64 & 133.64\\
\bottomrule[1.5pt]
\end{tabular}
\label{table_dataset}
\end{table}

\vspace{-0.1cm}
\subsection{Experiment Configurations}
To take full advantage of temporal information between consecutive video frames, in our experiments, we randomly choose $T=3$ consecutive frames from each real and fake video as the input of FTFDNet and FTFDNet-AVAM, and a further discussion on $T$ is conducted in ablation studies. Instead of the video frames, only the cropped face patches are input into the networks because of the occurrence of mouth shape forgery during talking face generation. The face detection is performed using S3FD \cite{34} and the obtained face crops are resized to $96 \times 96 \times 3$. The models are trained using Adam \cite{35} optimizer with default parameters ($\beta_1= 0.9$, $\beta_2= 0.999$), and initial learning rate equals to $0.001$.

\vspace{-0.1cm}
\subsection{Comparison Methods}
To demonstrate the advantages of the proposed audio-visual detection strategy, the comparative experiments were conducted as: (1) audio network: only use the audio stream in videos to detect fake talking faces, the network is composed of audio branch from FTFDNet and followed by three fully connected layers with dropout like FTFDNet. (2) visual network: same as audio network, but use visual branch from FTFDNet to replace the audio branch. (3) FTFDNet: our proposed audio-visual detection network. (4) FTFDNet-AVAM: FTFDNet integrated with the proposed audio-visual attention module. Furthermore, to reflect the superiority of our solution, the proposed models are also compared with a single-frame based model MesoInception-4 \cite{20}, which has similar model structure as ours. Noted that MesoInception-4 requires images of size $256 \times 256 \times 3$, in our experiment, face crops are resized to $256 \times 256 \times 3$ as its input.

\vspace{-0.1cm}
\subsection{Detection Results}

\subsubsection{Quantitative Analysis}
Two testing schemes are designed to verify the performance of our models: (1)Random: randomly sampling $T$ ($T=3$ for our models and $T=1$ for MesoInception-4) consecutive frames from each video in testing set (12936 videos altogether). (2) Definite: extracting $T$ consecutive images from the n-th ($n=10$ in experiments) frame of the video on testing set. All the detection results and the BCEloss (the lower the better) computed by Equation \ref{L_{BCE}} are shown in Table \ref{table_results}. Compared to MesoInception-4, the audio-only method has the least performance due to training un-convergence, while the visual-only method is much better owing to deeper network structure and the utilization of temporal context information between consecutive frames, but the limitation of resizing is hard to be ruled out. The performance of visual-only method to be inferior to the proposed audio-visual detection, which also confirms that the audio information can indeed benefit the FTFDNet. In particular, the addition of AVAM has substantially achieved the highest detection accuracy of 97.00\% and the lowest BCELoss of 0.0865.


\subsubsection{Qualitative Analysis}
\indent Fig. \ref{detection_results} shows the visualization of detection results of the best-performing detecting model (FTFDNet-AVAM). In Fig. \ref{detection_results}, the genuine faces are marked by a green outline and the generated fake faces generated are marked by a red outline. To reveal the influence of the talking face forgery, the first line presents the pristine face frames from the video, the next two lines present fake talking faces in testing set we established. By observing the enlarged mouth areas, it can be found that the fake talking faces are unidentifiable by the naked eyes, which also validates the advantage of the proposed high-precision detection model from another point of view.

\begin{table}[t]
\caption{\textbf{The quantitative results. }We test all models using two methods: Random, $T$ ($T=3$ for our models and $T=1$ for MesoInception-4) consecutive face images are randomly choose from each video. Definite, $T$ consecutive images are extracted from the 10th frame of the video on test set. Our FTFDNet-AVAM obtained the best performance.}
\vspace{-0.7cm}
$$
\setlength{\arraycolsep}{1mm}{
\begin{array}{lccccc}
\toprule[1.5pt]
\rule{0pt}{6pt}
 & \multicolumn{2}{c} {\text {    Random   }} & & \multicolumn{2}{c} {\text {   Definite   }} \\
\cline { 2 - 3 } \cline { 5 - 6 }
\rule{0pt}{10pt}
\text { Method } & \text {Acc}\uparrow & \text {BCELoss}\downarrow & & \text {Acc}\uparrow & \text {BCELoss}\downarrow \\
\hline
\rule{0pt}{10pt}
\textbf {MesoInception-4\cite{20}} & 85.76 & 0.4005 & & 86.07 & 0.3743\\
\textbf {Audio(ours)} & 59.89 & 0.6643 & & 59.52 & 0.6655\\
\textbf {Visual(ours)} & 96.30 & 0.1011 & & 96.24 & 0.1092\\
\textbf {FTFDNet(ours)} & 96.56 & 0.0933 & & 96.54 & 0.0938\\
\textbf {FTFDNet-AVAM(ours)} & \textbf{97.00} & \textbf{0.0865} & & \textbf{97.01} & \textbf{0.0792}\\
\bottomrule[1.5pt]
\end{array}
}
$$
\label{table_results}
\end{table}

\subsection{Ablation Studies}
\subsubsection{Ablation Studies on $T$}
\cite{22} indicated that the temporal information between consecutive video frames is essential for DeepFake detection because of the inter-frame tiny artifacts. For the talking face generation model Wav2Lip \cite{26}, a buffer of $T$ ($T=5$ as originally defined) frames is used to obtain the temporal context information in lip-sync. More than these inspirations, the talking face generation guided by the given speech also leads to a strong correlation between lips from consecutive video frames.

In our experiments, the ablation study is conducted on the buffered consecutive frames, which are randomly-sampled from the video on FTFDD as the input of networks. Table \ref{Table_T} presents the detection rate and BCELoss of $T=1$, $T=3$, $T=5$ on both visual-only network and audio-visual network (FTFDNet). It can be found that the inter-frame information has indeed enhanced the accuracy of fake talking face detection not only for visual-only network but also for FTFDNet (The accuracy of $T=3$ is higher than the accuracy of $T=1$) as expected. At the same time, it is found that a bigger buffer size does not mean better performance (The accuracy of $T=5$ is lower than the accuracy of $T=3$). One possible reason is that the accumulated temporal context information between the frames does also increase the complexity of the recognition/classification, which needs to be balanced.

\begin{table}[t]
\caption{\textbf{The ablation studies on T. }It's observed that when $T=3$,  both visual-only network and audio-visual network (FTFDNet) perform the best.}
\vspace{-0.5cm}
$$
\setlength{\arraycolsep}{1mm}{
\begin{array}{lccccc}
\toprule[1.5pt]
\rule{0pt}{6pt}
 & \multicolumn{2}{c} {\text {    Visual   }} & & \multicolumn{2}{c} {\text {   Audio-Visual   }} \\
\cline { 2 - 3 } \cline { 5 - 6 }
\rule{0pt}{10pt}
\text { T } & \text {Acc}\uparrow & \text {BCELoss}\downarrow & & \text {Acc}\uparrow & \text {BCELoss}\downarrow \\
\hline
\rule{0pt}{10pt}
\textbf {T = 1} & 95.44 &0.1390 & & 96.27 & 0.1046\\
\textbf {T = 3} & \textbf{96.30} & \textbf{0.1011} & & \textbf{96.56} & \textbf{0.0933}\\
\textbf {T = 5} & 95.59 & 0.1293 & & 96.50 & 0.1033\\
\bottomrule[1.5pt]
\end{array}
}
$$
\label{Table_T}
\end{table}

\begin{table}[t]
\caption{\textbf{The ablation studies on AVAM. }To verify the validity of AVAM, we compared it with CBAM, and found that audio-visual fusion does improve the performance of attention mechanism.}
\vspace{-0.5cm}
$$
\setlength{\arraycolsep}{1mm}{
\begin{array}{lccccc}
\toprule[1.5pt]
\rule{0pt}{6pt}
 & \multicolumn{2}{c} {\text {    Random   }} & & \multicolumn{2}{c} {\text {   Definite   }} \\
\cline { 2 - 3 } \cline { 5 - 6 }
\rule{0pt}{10pt}
\text { Network } & \text {Acc}\uparrow & \text {BCELoss}\downarrow & & \text {Acc}\uparrow & \text {BCELoss}\downarrow \\
\hline
\rule{0pt}{10pt}
\textbf {FTFDNet} & 96.56 & 0.0933 & & 96.54 & 0.0938\\
\textbf {FTFDNet-CBAM} & 96.73 & 0.099 & & 96.75 & 0.0904\\
\textbf {FTFDNet-AVAM} & \textbf{97.00} & \textbf{0.0865} & & \textbf{97.01} & \textbf{0.0792}\\
\bottomrule[1.5pt]
\end{array}
}
$$
\label{Table_AVAM}
\end{table}

\subsubsection{Ablation Studies on AVAM}
To compare the proposed audio-visual attention mechanism with the conventional visual attention mechanism, we incorporate the CBAM\cite{30} into FTFDNet as FTFD-CBAM in a similar way as FTFDNet-AVAM. Unlike CBAM, our AVAM only explores inter-spatial relationship of features like the spacial attention module (SAM) in CBAM, but excludes the inter-channel relationship of features as the channel attention module (CAM) in CBAM. Through experiments, it is found that the CAM has very limited contribution to the detection improvement. Therefore, the CBAM incorporated in our model is only composed of spacial attention module. As shown in Table \ref{Table_AVAM}, not only CBAM but also AVAM can improve the performance of detection model, and our FTFDNet-AVAM achieves detection accuracy of 97.00\% and BCELoss of 0.0865, which is superior to the performance of  FTFDNet-CBAM with detection accuracy of 96.73\% and BCELoss of 0.099. The experimental results have demonstrated that the audio information is useful to attention mechanism in audio-visual tasks.

\begin{figure}[t]
\centering
\centerline{\includegraphics [width=0.3\textwidth]{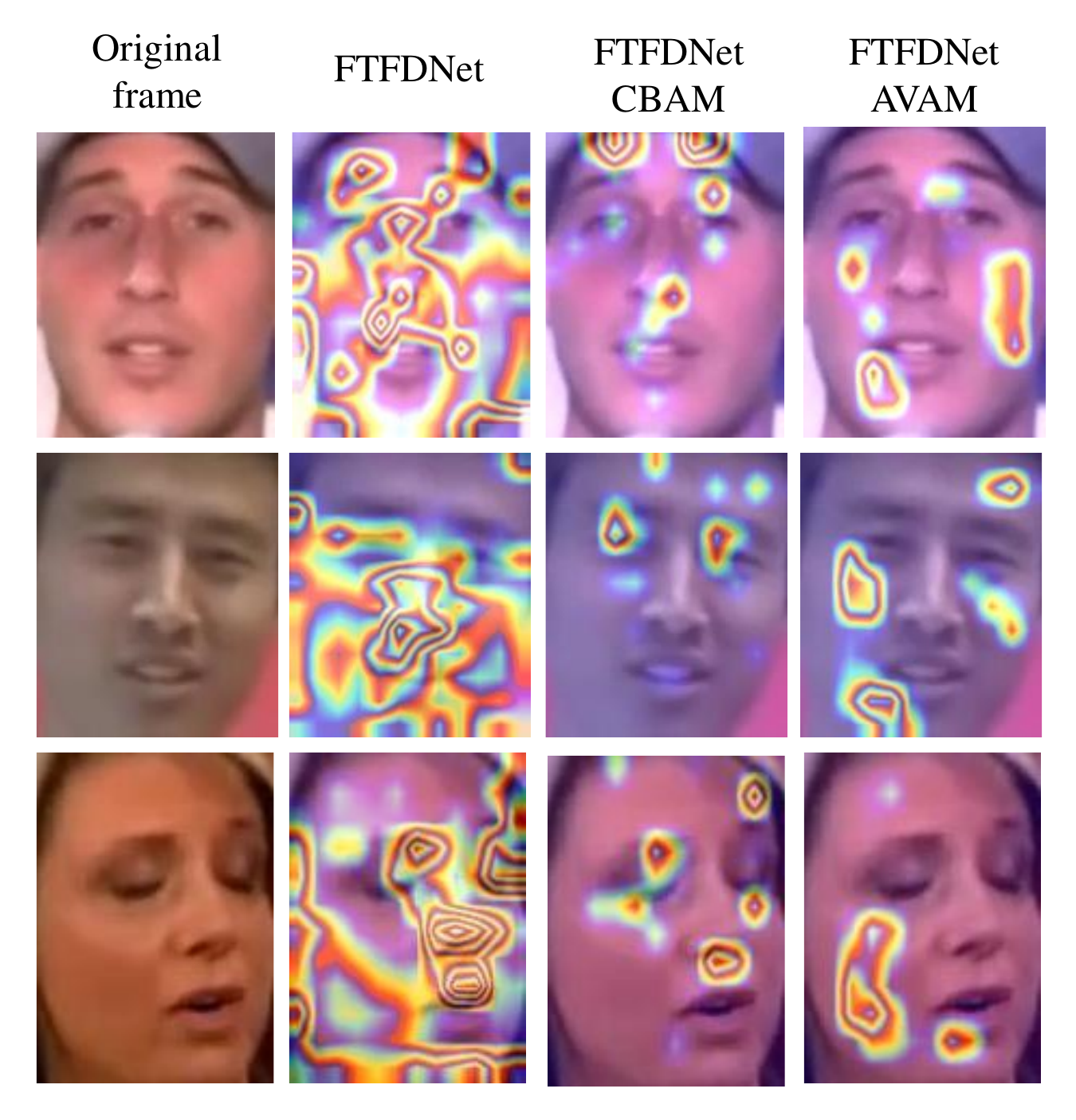}}
\setlength{\abovecaptionskip}{0.cm}
\caption{\textbf{Visualization of attentional feature map. }Some examples of intermediate feature maps from FTFDNet, FTFDNet-CBAM and FTFDNet-AVAM are visualized in their original face crops.}
\label{attention map}
\vspace{-0.2cm}
\end{figure}

Fig. \ref{attention map} shows the visualized attentional maps, where the feature maps are also visualized from the visual branch of FTFDNet, FTFDNet-CBAM and FTFDNet-AVAM. It is found that after adding the attention mechanism, the network produces different degrees of attention to different regions in the whole face image. Compared to the conventional visual attention CBAM, the proposed AVAM enables the fake detection network pay more attention to the lip and cheek regions which might be tampered by talking face generation methods. This verifies the usefulness of audio information for key regions extraction. Once these critical regions have been emphasized, the fake talking face detection network is able to achieve higher detection accuracy.

\vspace{-0.2cm}
\section{Conclusion}
In this paper, we propose a novel fake talking face detection network (FTFDNet) by incorporating audio and visual representation, which is inspired by the decision-making mechanism of human multisensory perception system. Beyond that, we propose an audio-visual attention mechanism (AVAM), which enables the network to focus only on the most relevant portions of the feature maps. To further improve the performance of the detection of fake talking face frames from videos, our proposed AVAM was also embeded into FTFDNet, called FTFDNet-AVAM. Training and evaluating of the proposed works are performed on an newly established dataset (FTFDD), and the proposed FTFDNet-AVAM shows an excellent performance on the detection of fake talking face videos with an detection rate of 97\%.

\section*{Acknowledgments}
This work was supported in part by the National Natural Science Foundation of China under Grants 61971352 and 61862043, in part by the Natural Science Foundation of Jiangxi Province under Grant 20204BCJ22011.

\bibliographystyle{IEEEtran}
\bibliography{inference}

\vfill

\end{document}